\documentclass[sigconf]{acmart}
\renewcommand\footnotetextcopyrightpermission[1]{}

\usepackage{subcaption}
\usepackage{xcolor}
\definecolor{plusgreen}{RGB}{43,150,20}
\usepackage{hyperref}
\hypersetup{
    colorlinks=true,
    linkcolor=blue,
    filecolor=magenta,      
    urlcolor=cyan
}
\usepackage{float}
\usepackage{multirow}
\usepackage{graphicx}
\usepackage{fancyhdr}

\begin{document}

\pagestyle{fancy}
\fancyhead{}

\title[Medical Face Masks and Emotion Recognition from the Body]{Medical Face Masks and Emotion Recognition from the Body: Insights from a Deep Learning Perspective}


\author{Nikolaos Kegkeroglou}
\affiliation{%
\institution{National Technical University of Athens, School of ECE}
\city{15773 Athens}
\country{Greece}}
\email{nkegkeroglou@gmail.com}

\author{Panagiotis P. Filntisis}
\affiliation{%
\institution{Athena Research Center}
\department{Institute of Robotics}
\city{15125 Maroussi}
\country{Greece}}
\email{filby@central.ntua.gr}

\author{Petros Maragos}
\affiliation{%
\institution{National Technical University of Athens, School of ECE}
\city{15773 Athens}
\country{Greece}}
\affiliation{%
\institution{Athena Research Center}
\department{Institute of Robotics}
\city{15125 Maroussi}
\country{Greece}}
\email{maragos@cs.ntua.gr}

\begin{abstract}
The COVID-19 pandemic has undoubtedly changed the standards and affected all aspects of our lives, especially social communication. It has forced people to extensively wear medical face masks, in order to prevent transmission. This face occlusion can strongly irritate emotional reading from the face and urges us to incorporate the whole body as an emotional cue. In this paper, we conduct insightful studies about the effect of face occlusion on emotion recognition performance, and showcase the superiority of full body input over the plain masked face. We utilize a deep learning model based on the Temporal Segment Network framework, and aspire to fully overcome the face mask consequences. Although facial and bodily features can be learned from a single input, this may lead to irrelevant information confusion. By processing those features separately and fusing their prediction scores, we are more effectively taking advantage of both modalities. This framework also naturally supports temporal modeling, by mingling information among neighboring frames. In combination, these techniques form an effective system capable of tackling emotion recognition difficulties, caused by safety protocols applied in crucial areas.

\end{abstract}




\maketitle

\section{Introduction}
The possible applications of an interface capable of assessing human emotional states are numerous. Humans generally treat computer agents as they might treat other people \cite{RN96}. Robots and systems that are able to recognize, interpret and process human affect \cite{Br03}, are arguably well suited to this, making the interaction more effective and pleasant. They find fertile ground in the area of computer-assisted education, as learning is the quintessential emotional experience. A learning episode might begin with curiosity and fascination. But as its difficulty increases, one may experience confusion, frustration or anxiety, and thus, may abandon learning \cite{Pek02}. A tutoring agent, who is able to estimate the learner's affective state, can respond appropriately and give encouraging suggestions. Existing work has shown that robot tutors enhance learning, by personalizing their motivational strategies to the student's emotional behavior \cite{Gor16} \cite{Lei12}. Another crucial area is health care, as mental health disorders, like depression and psychosis, are on the rise across the world. Emotion recognition systems can be an effective strategy for preventing and monitoring such disorders \cite{Yang21}.

While works based on facial expressions abound in the area, recognizing affect from the body remains a less explored topic. A study in neurobiology has shown that body movement and posture contain useful features for recognizing human affect \cite{deG09}. In other experiments, it was shown that facial and bodily expressions work complementary for visual perception of emotion, and in some cases humans perceive bodily expressed emotional information as more diagnostic than facial \cite{Avi12}. Furthermore, the visibility of facial cues is not guaranteed. Bodily expression recognition becomes crucial when facial features are occluded. Medical face masks, which are extensively used nowadays due to the COVID-19 pandemic \cite{Car20}, are the epitome of face occlusion. Because bodies are more expressive than faces in those situations, social information can be detected from the body instead.

Although there has been a considerable amount of research on automatic emotion recognition in adults, the topic regarding children has been understudied. Children go through a critical development process and applications involving them require special attention \cite{Re13}. They also tend to fidget and move around more than adults, leading to more self-occlusions and non-frontal head poses \cite{Bel13}. This becomes even more challenging, considering the current health and safety protocols that demand the use of face masks. Robots can no longer rely only on facial expressions to recognize emotion, but also have to take into account body expressions that can stay visible and detectable, even when the face is unobservable. Children’s behavior and natural characteristics differ from adults, so perception systems need to be specifically trained, to be able to tackle Child-Robot Interaction (CRI) problems.

The rest of the paper is organized as follows: Section \ref{section:work} discusses related work on emotion recognition mainly from the body. Section \ref{section:method} describes the adopted deep learning-based visual emotion recognition model in detail, as well as the tools and methods used for the experiments. Section \ref{section:res} presents the experimental results, and lastly Section \ref{section:conc} provides our conclusions.

\section{Related Work}
\label{section:work}
In recent years, deep learning methods have been very popular due to the massive amounts of digital data in combination with powerful processing hardware. Deep extracted features have yielded excellent results and on most cases outperformed non-deep state-of-the-art methods for the emotion recognition task. When processing a video with emotional expressions, an essential component is capturing temporal information to complement the prediction from still images. Two-stream Convolutional Neural Network (CNN) architectures use multi-frame optical flow to handle complex actions like emotional expressions \cite{KZ14}.

The most common modality used by the research community for identifying emotion is facial expressions \cite{LiDe20}. Some works have proposed an audiovisual approach \cite{FEmo}, where the system takes speech as an additional input to the face, in order to tackle occlusions and increase robustness. In \cite{C3D}, they utilize 3D CNNs to extract spatio-temporal features both from face videos and audio signals, and deep belief nets \cite{DBN} for emotion recognition. However, the COVID-19 pandemic has fostered a pervasive use of medical face masks all around the world, making a serious impact on social communication. Several studies investigated how the presence of a face mask affects emotion recognition accuracy and revealed that it diminishes the people’s ability to accurately categorize a facial expression \cite{Car20} \cite{mask1}. On top of that, the mask impairs re-identification of the same face by people \cite{mask3}, which suggests a need for mask-specific model training. In \cite{mask2}, they also explored how masks influence the perceived emotional profile of facial expressions. It was shown, that it not only led to a decrease in perceived intensity of the intended emotions, but it also resulted in an overall increase in the perceived intensity of non-intended emotions. In \cite{mask4}, even super-recognizers, people who are highly skilled and superior in recognition tasks, were impaired by the face occlusion caused by the face mask. This negative effect in emotional reading is not limited to adults, as it also concerns interaction with children \cite{mask5}.

Motivated by all the above, we move towards incorporating bodily expressed information as a major cue in the emotion recognition task. An early work \cite{GP06} combined handcrafted face and body features at feature and decision-level for emotion classification. In \cite{F19}, a hierarchical multi-label annotation method was proposed, which fused body skeleton with facial expressions for automatic recognition of emotion of children during CRI scenarios. In \cite{Luo+20}, they experimented with two bodily expression pipelines, one of which implemented a two-stream-based CNN. The other one relied solely on the human skeleton and utilized a spatial-temporal Graph Convolutional Network (GCN) \cite{Yan+18}, which constructs a graph from the human body landmarks with their natural spatial connectivity, as well as temporally neighboring landmarks. 

Along with body, context has been an additional modality involved in the task of emotion recognition. In \cite{F20}, RGB and flow body streams were accompanied with a context RGB stream and a visual-semantic embedding loss based on word embedding representations. In \cite{Huang}, they proposed a network structure composed of a GCN processing skeleton landmarks, and two 3D CNNs for RGB body and context input. A network ensemble, including streams that processed the body in RGB, flow and skeleton form was proposed in \cite{Pik+21}. This variety of bodily expressed cues have also been involved in CRI emotion recognition systems \cite{N21} \cite{marin}.

Our work focuses on the effect of the face occlusion on emotion recognition performance. We adopt a proven related work model and process only RGB input, despite the diversity of body cues that can be conveyed. We sense that this approach suits best to our purpose, regarding the medical face mask effect study.

\begin{figure*}[th!]
    \centering
        \includegraphics[width=.8\linewidth]{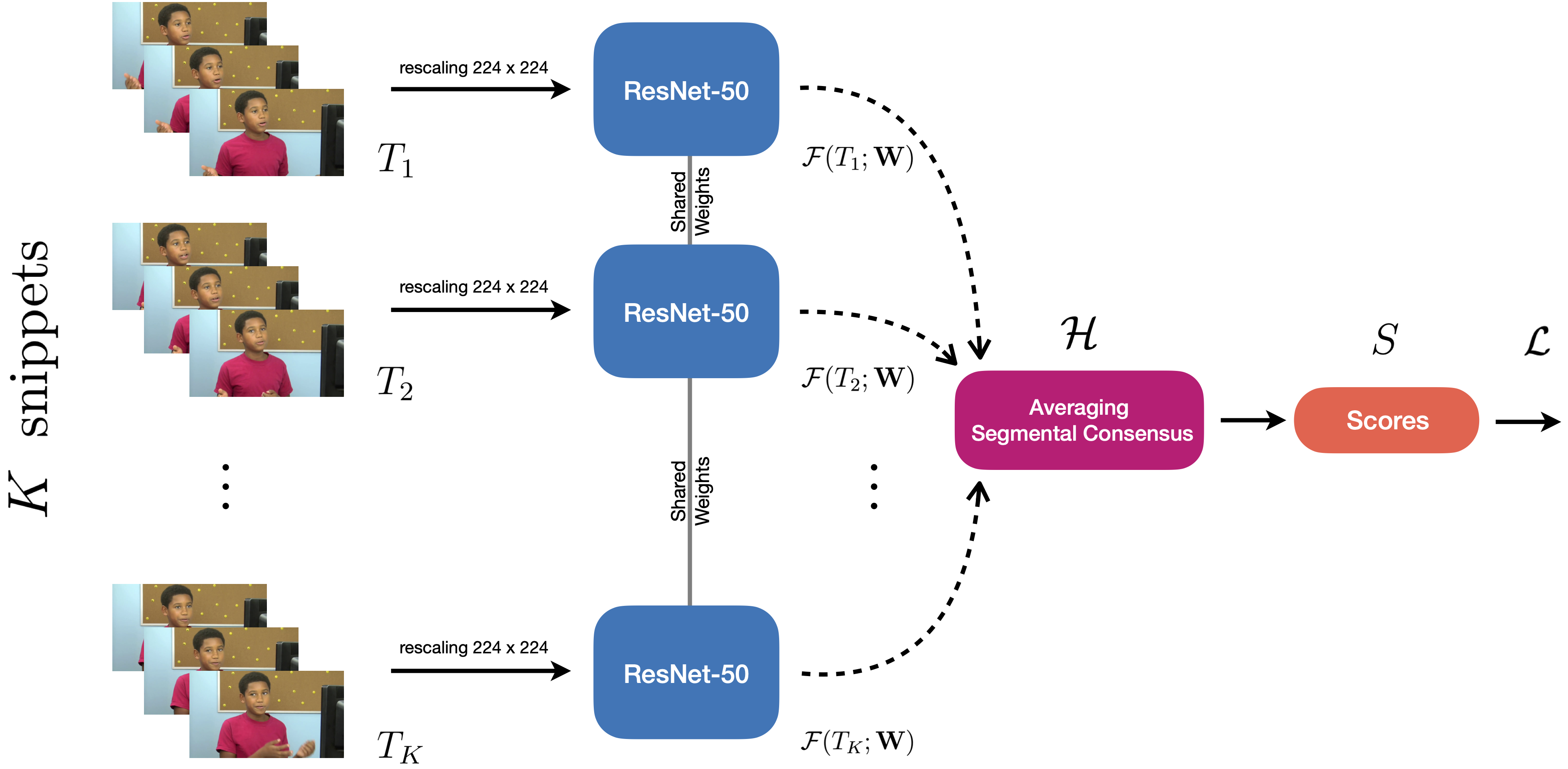}
    \caption{TSN-Based Model Architecture}
    \label{baselinemodel}
\end{figure*}

\section{Visual Emotion Recognition Model}
\label{section:method}
In this chapter, we present the model, that will be used to tackle the visual emotion recognition task. We discuss its structure and benefits and also address the occuring challenges, which are taken into account in the model's various configurations. Furthermore, we describe the tools utilized to conduct the upcoming experiments and some techniques to enhance model performance.

    \subsection{Feature Capturing}
    Complex actions, like emotional expressions, comprise multiple stages spanning over a period of time and it would be quite a loss failing to utilize them. On the other hand, each expressed emotion is not present throughout a whole input video. These facts, indicate that we are in need of effective general feature capturing. While the plain CNN architecture considers the whole input sequence, as well as each frame in the video separately, the Temporal Segment Network (TSN) framework \cite{Wa+16} operates on a sequence of short snippets sparsely sampled from the entire video. Each snippet in this sequence will produce its own preliminary prediction of the emotion classes and then, a consensus among the snippets will be derived as the video-level prediction. Therefore, it allows the network to access several parts of the video, but also tackles the inability of the former to model long-range temporal structure, thus, being more likely to observe the corresponding expression.

    \subsection{Method}
    The overall architecture of our model is shown in Fig. \ref{baselinemodel}. Formally, given a video $V$, we divide it into $K$ non-overlapping segments $\{S_{1}, S_{2}, ..., S_{K}\}$, to access several parts of the video, and transform them into a sequence of snippets $\{T_{1}, T_{2}, ... , T_{K}\}$. Each snippet $T_{k}$ is produced, by randomly sampling 3 consecutive frames from its corresponding segment $S_{k}$, to tackle frame redundancy. Finally, a segmental consensus function $\mathcal{H}$ is applied on the snippet-level predictions produced by the backbone, to obtain the final scores $S$:
    $$S = \text{TSN}(T_{1}, T_{2}, ..., T_{K}) = \mathcal{H}(\mathcal{F}(T_{1};\textbf{W}), \mathcal{F}(T_{2};\textbf{W}), ... , \mathcal{F}(T_{K};\textbf{W}))$$
    Here $\mathcal{F}(T_{k};\textbf{W})$ denotes the function representing the application of a CNN with parameters $\textbf{W}$ on the snippet $T_{k}$. The CNN is equipped with a ResNet-50 backbone architecture \cite{He+16}. The consensus function $\mathcal{H}$ we use is average pooling and the obtained video-level scores $S$ are fed to a loss function $\mathcal{L}$ to perform the training step. This framework offers several benefits to emotion recognition. Compared to processing the entire video, the sampling process ignores redundant information in consecutive video frames, helping avoid overfitting, and offers a type of data augmentation, valuable for children emotion databases of small size.

    \begin{figure}[t!]
        \centering
        \includegraphics[width=.995\linewidth]{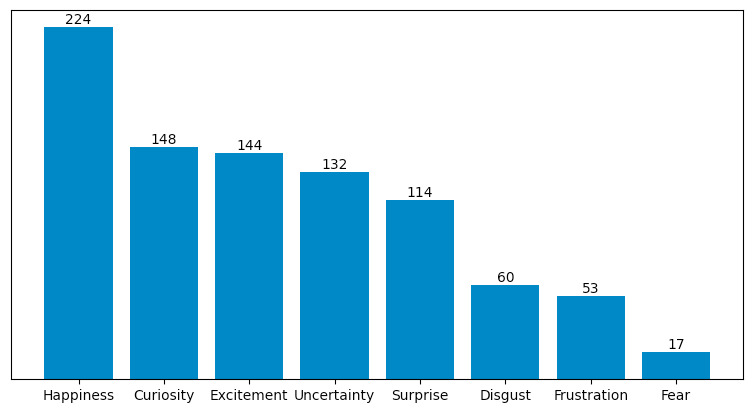}
        \caption{EmoReact Training Set Imbalance}
        \label{emoimb}
    \end{figure}

    \subsection{Database}
    We perform our experiments on the EmoReact dataset \cite{No+16}, which contains 1102 videos of 63 children, aged between 4 and 14, expressing emotions while discussing different topics, collected from the YouTube channel React. Each video is annotated with one or more emotions, from a total of 8 emotion labels: Curiosity, Uncertainty, Excitement, Happiness, Surprise, Disgust, Fear, and Frustration. Therefore, we are dealing with a CRI binary multi-label classification problem. In Fig. \ref{emoimb}, we show the imbalance of EmoReact's training set, which means it includes an unequal number of videos for each emotion label, and is something that we must address in our upcoming model configuration choices. We can also argue that some emotions (Fear, Frustration, Disgust) are expressed in a relatively low number of samples, which results in possible lack of diversity and less ease to generalize well across unseen individuals, introducing an extra degree of difficulty to our problem.

    \subsection{Medical Face Mask Effect Study}
    The COVID-19 pandemic has forced people to extensively wear medical face masks, in order to prevent transmission. Motivated by this fact, we want to conduct an experimental study about the effect of medical face masks on emotion recognition, by applying a relevant mask on the EmoReact children's faces, as an attempt to simulate the face occlusion consequence.

            \begin{figure}[t!]
			\centering
			\begin{subfigure}[b]{.32\linewidth}
   				\includegraphics[width=1\linewidth]{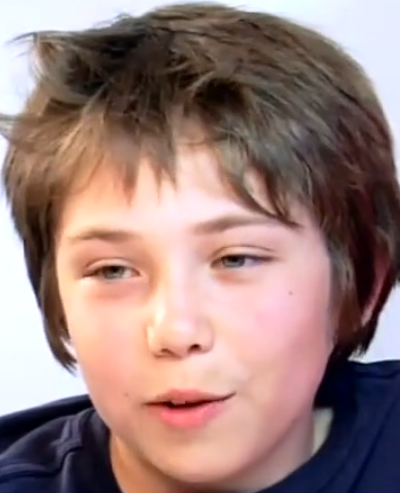}
   				\caption{Original Image}
   				\label{maska} 
			\end{subfigure}
			\begin{subfigure}[b]{.32\linewidth}
   				\includegraphics[width=1\linewidth]{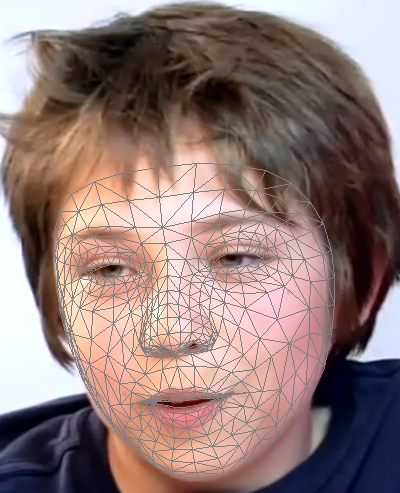}
   				\caption{Mesh Tracking}
   				\label{maskb} 
			\end{subfigure}
			\begin{subfigure}[b]{.32\linewidth}
   				\includegraphics[width=1\linewidth]{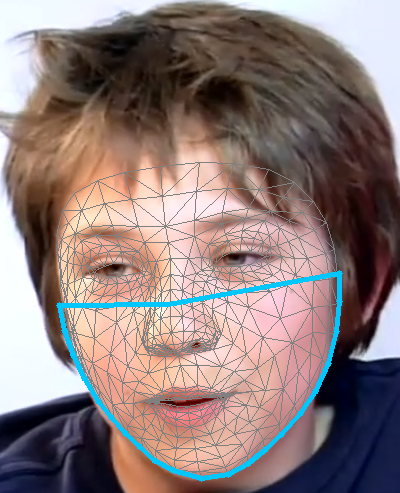}
   				\caption{Mask Polygon}
   				\label{maskc} 
			\end{subfigure}
			\caption{Mask Application Steps}
			\label{mask}
		\end{figure}
  
    \subsubsection{Mask Application}
\begin{figure}[t!]
			\centering
   				\includegraphics[width=.4775\linewidth]{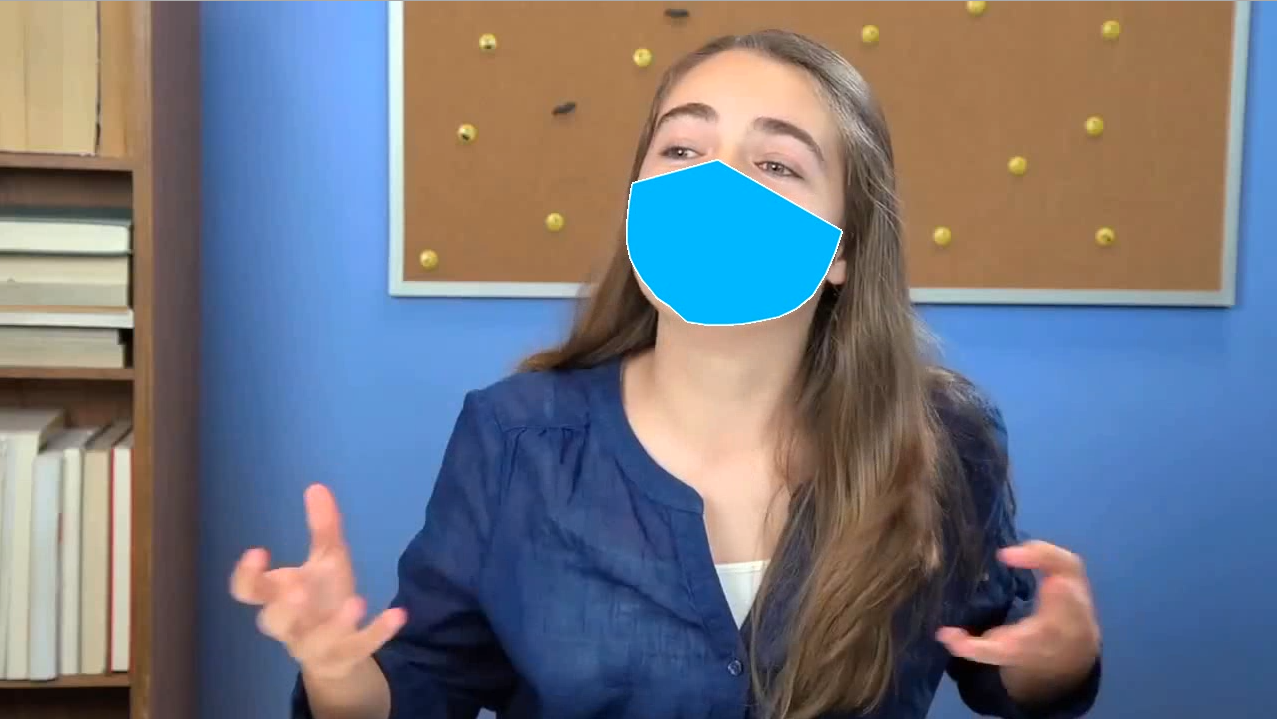}
   				\includegraphics[width=.4775\linewidth]{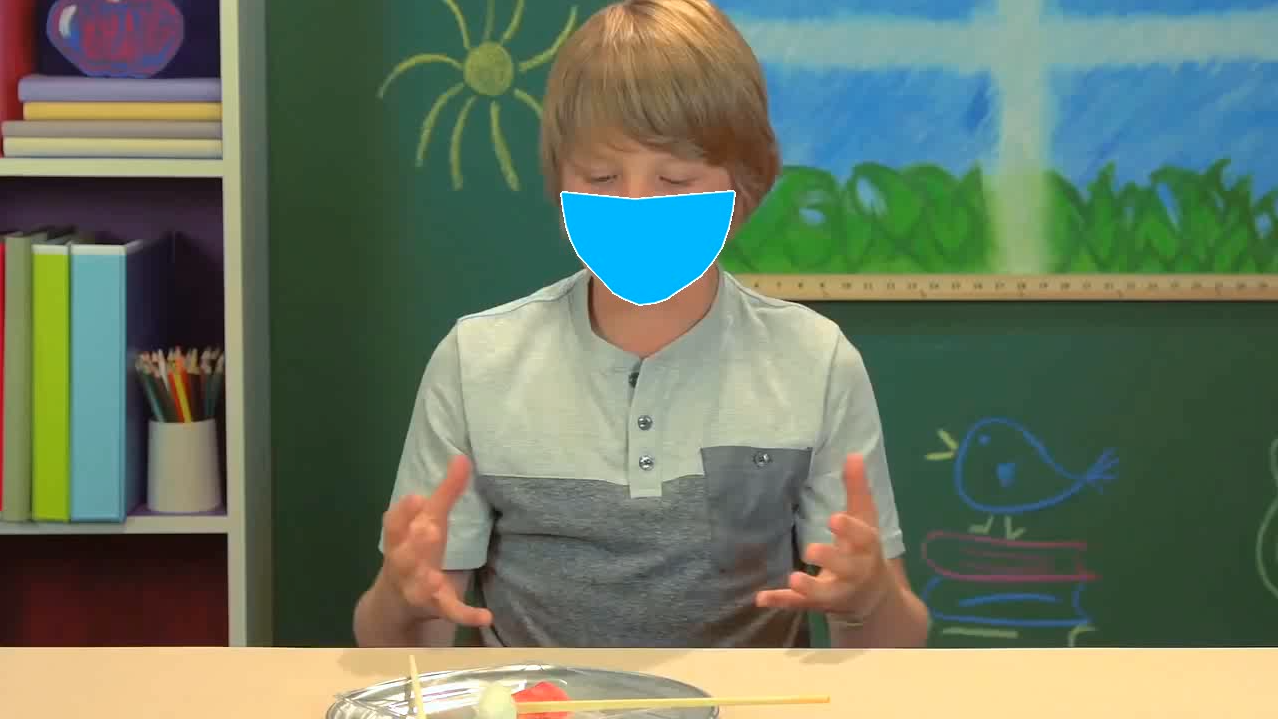}
   				\includegraphics[width=.24125\linewidth]{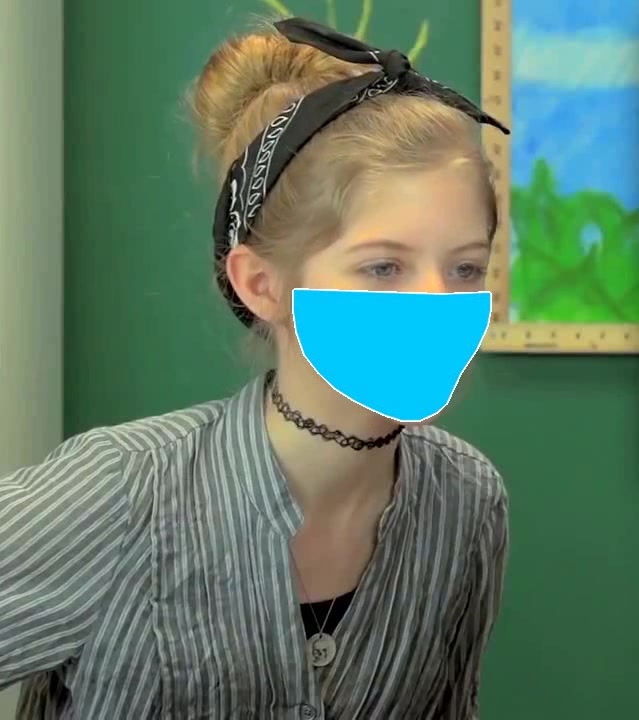}
   				\includegraphics[width=.445\linewidth]{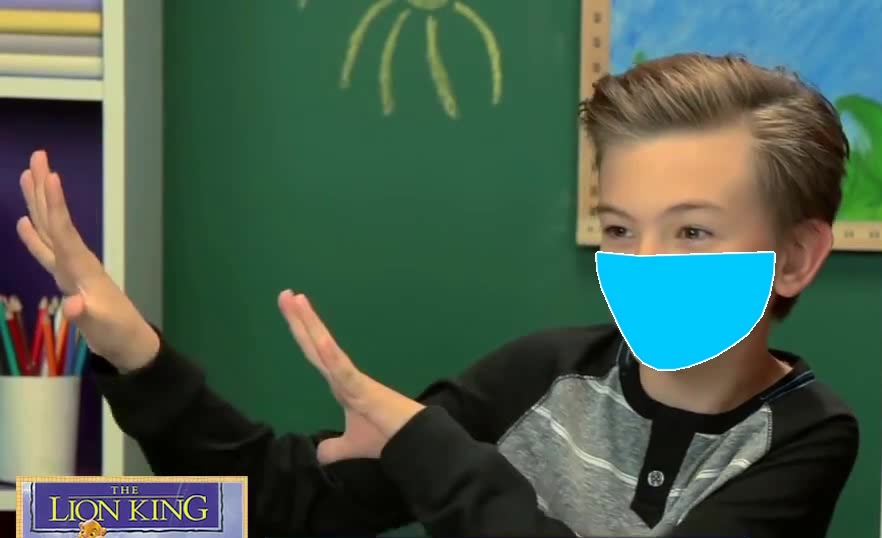}
   				\includegraphics[width=.25875\linewidth]{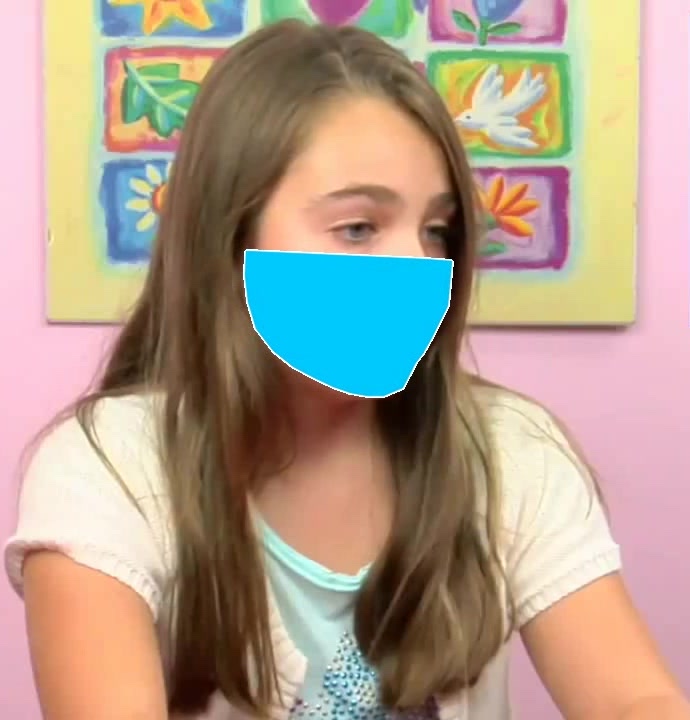}
			\caption{EmoReact Masked Samples}
			\label{emomask}
		\end{figure}

    To apply the mask, we detect the facial surface geometry using Google's MediaPipe Face Mesh \cite{mediapipe}, an end-to-end CNN-based model for inferring an approximate 3D mesh representation of a human face from a single image. It uses a dense mesh model of 468 vertices and is well-suited for face-based augmented reality effects. We track the 2D coordinates of the right and left jawline vertices, starting from just below the eyes until the chin, and one extra vertex for the nose, in order to form a polygon that is finally filled to represent the mask (Fig. \ref{mask}). The jawlines for the mask are created by tracking the edge x-axis vertices and accordingly selecting among several jawline candidates, that we manually created for this particular face mesh model \footnote{The code for the mask application tool is publicly available at: \href{https://github.com/nkegke/medical-face-mask-applier}{https://github.com/nkegke/medical-face-mask-applier}}. In Fig. \ref{emomask}, we display several samples of EmoReact after the application of the mask and showcase our tool's robustness to face orientation.

    \subsubsection{Body Detection}
\begin{figure}[t!]
        \centering
            \includegraphics[width=1\linewidth]{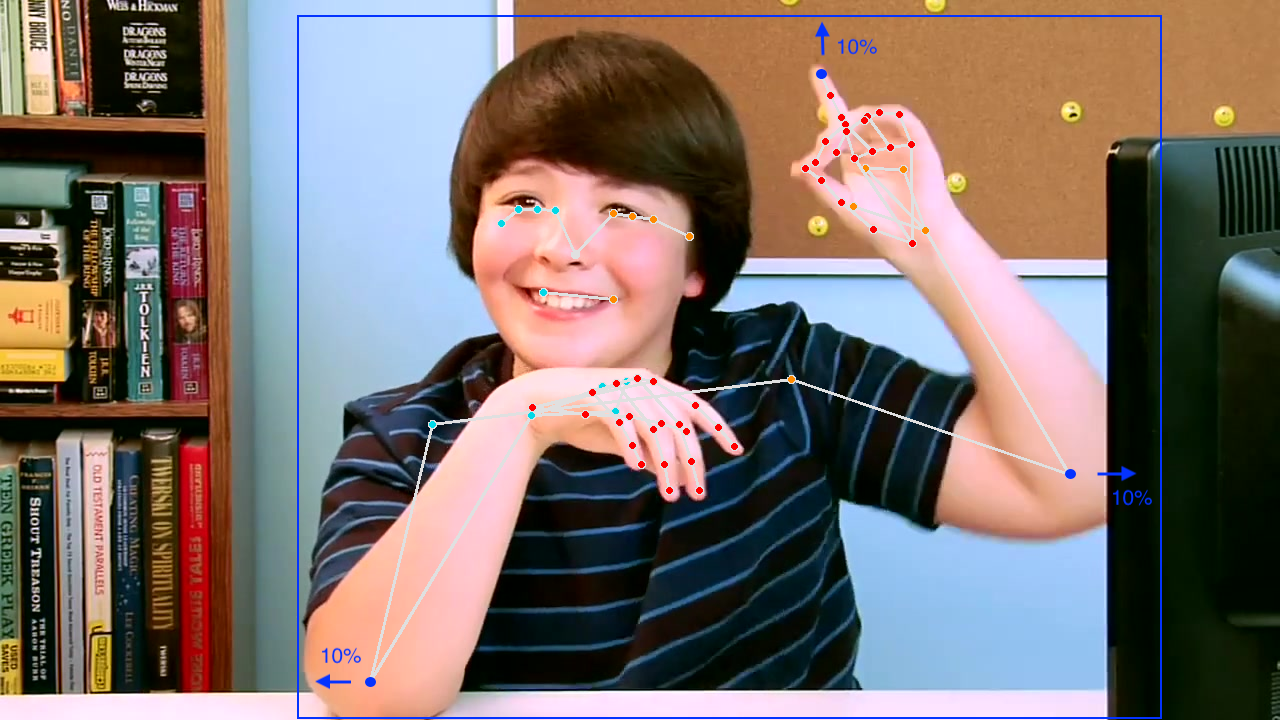}
            \caption{Body Detection}
            \label{body}
    \end{figure}

    In order to incorporate bodily expressions, we need a way to detect the human body. Google's MediaPipe also provides human body and hand skeleton tracking tools \cite{blazepose} \cite{mphands}. We combine keypoints tracked by both tools and create a bounding box with the edge points, expanded by a factor of 10\% at each respective dimension, which is then cropped as the input image. This process is demonstrated in Fig. \ref{body}, where most background noise is removed and full body information dominates the cropped image.

    \subsection{Modality Fusion}
\begin{figure*}[th!]
        \centering
        \includegraphics[width=.85\linewidth]{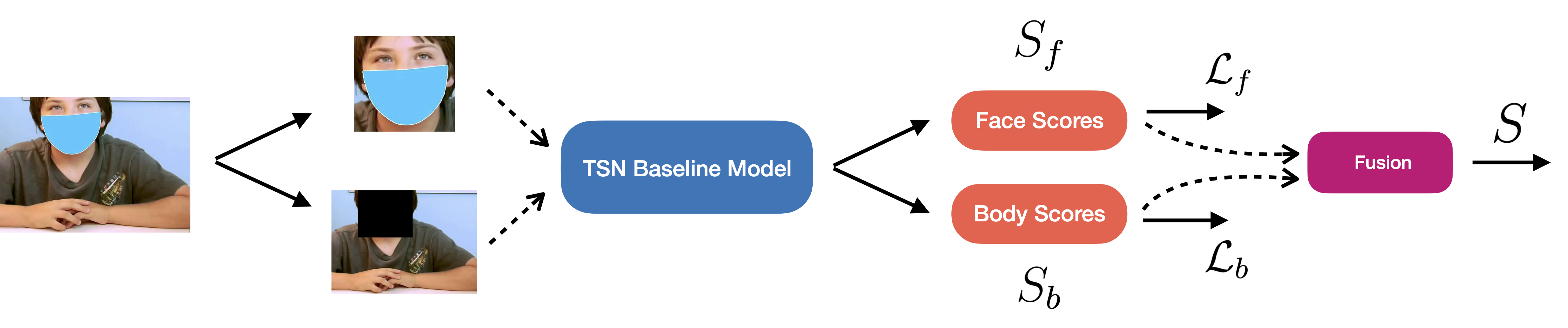}
        \caption{Modality Score Late Fusion Scheme}
        \label{fusion}
    \end{figure*}

    We are looking to take advantage of the face and body information separately, by fusing the individual modality prediction scores with a late fusion scheme. The full body crop includes the masked face, and processing it as a single RGB input image can lead to irrelevant information confusion. The proposed method is to separate the face and body features, in order to avoid the aforementioned issue. The core model remains as is, but now processes the face crop, and the plain body crop with the corresponding face area blacked out, in two separate forward passes (Fig. \ref{fusion}). After producing the scores $S_{f}$ and $S_{b}$ from face and plain body respectively, we use a late fusion scheme to obtain the final scores $S$. Finally, the overall loss $\mathcal{L}$ is simply the summation of the individual modality losses:  $\mathcal{L} = L_{f} + L_{b}$.

    \subsection{Temporal Modeling}
    The current TSN-based model processes only one of the $N$ consecutive frames of each snippet, being heavily based on spatial structure. This architecture naturally supports temporal modeling, by mingling information among neighboring snippet frames with the Temporal Shift Module (TSM) \cite{Lin+19}. TSM can be inserted into CNNs, to exploit temporality at zero computation and parameters. It shifts part of the channels of the input frames and the latent representations of each snippet along the temporal dimension, both forward and backward, thus facilitate information exchange among neighboring frames. Because information contained in the channels is no longer accessible for the current frame, the spatial modeling ability of the backbone can be harmed. To address this problem, the module is placed inside the residual branches of the ResNet, so the information in the original activation is still accessible after temporal shift, through the identity mappings.

    \subsection{Model \& Training Configurations}
    The model is pretrained on AffectNet, the largest facial expression dataset. We obtain the weights of the network as provided by the PyTorch framework, achieving 59.47\% accuracy on the validation set. Before feeding the input to the network, we rescale sampled RGB images from full resolution to 224 $\times$ 224. We train our models for 60 epochs, with stochastic gradient descent with momentum 0.9 and a batch size of 8, L2 regularization with weight decay 5e-4, and start with a learning rate of 1e-2, which is then reduced by a factor of 10 at 20 and 40 epoch milestones\footnote{The code for the model and the experiments is publicly available at: \href{https://github.com/nkegke/deep-affective-bodily-expression-recognition}{https://github.com/nkegke/deep-affective-bodily-expression-recognition}}. Since our task is binary multi-label classification, our predictions are fed to a binary cross-entropy (BCE) loss function, after suppressing the scores $S$ to [0,1] with a sigmoid function. BCE depends on the the label-specific error, thus it penalizes label predictions independently. Following prior work, the only evaluation metric that has been shown to be robust to imbalanced datasets is the Area Under the Curve of Receiver Operating Characteristic (ROC AUC). The scores $S$, of size 8 per sample, are averaged to obtain a single overall performance metric. Instead of giving equal weight to each class, which will over-emphasize on the typically low performance on an infrequent class, we compute the unbalanced average, so every sample-class pair contributes equally to the overall metric. For evaluation, we select the epoch with the best validation ROC AUC and apply the corresponding network on the test set, to finally report the best overall performance achieved.
    
\section{Experimental Results}
\label{section:res}
In this section, we present our experimental procedure and results. First, an ablation study on the number of TSN segments is performed to explore possible trade-offs. Then, we study the medical face mask effect by comparing emotion recognition results of masked input to when the faces are visible. We examine the case of when the mask is applied to the image of the full body, as well as only the face, to compare performance between input modalities. Furthermore, visual explanation techniques are utilized to display expressive features for different modalities and emotion categories. Lastly, we report results given with the enhancement techniques, both when individually utilized and when combined.

    \subsection{Performance vs Speed Trade-off}
    In Table \ref{loadtab}, we perform an ablation study on the number of segments and consequently the number of snippets, which are used during the TSN training, by considering 4 different values: 1, 3, 5 and 10. By increasing the number of segments, we significantly increase computational load, and therefore inference and training time. On the other hand, when we provide the model with multiple parts of the video, it might help achieve better performance. The numbers reported stand for training with a single RTX 2080 GPU, but one could use multiple ones and increase batch size proportionally for faster training.

    \begin{table}[t!]
    \centering
    \caption{TSN Training Computational Load}
    \begin{tabular}{|c|cc|}
    \hline
    \multirow{2}{*}{\textbf{Segments}} & \multicolumn{2}{c|}{\textbf{Time per Epoch (sec.)}} \\ \cline{2-3}
                                       & \multicolumn{1}{c|}{Training} & Validation \\ \hline
    1  & \multicolumn{1}{c|}{6}     & 4 \\ \hline
    3  & \multicolumn{1}{c|}{14}    & 10 \\ \hline
    5  & \multicolumn{1}{c|}{23}    & 16 \\ \hline
    10 & \multicolumn{1}{c|}{32}    & 23 \\ \hline                  
    \end{tabular}
    \label{loadtab}
    \end{table}
        
    \subsection{Mask Effect Results}
    We compare emotion recognition results between default and masked input, for face and full body crops. For face cropping, we extract the visual face features using OpenFace \cite{openface}, an open source facial behavior analysis toolkit.
    
        \subsubsection{Mask Effect on Face Input}
        In Table \ref{effect1}, we report results on face input. At first sight, performance drops considerably ($\approx$ 3-4\%). This is a result we expected, as the mask covers the majority of the face, including one of the most expressive facial features, the mouth. Intuitively, if one would try to predict the emotions expressed in the two images of Table \ref{effect1}, we sense that they would have a better chance without the presence of the mask. Regarding the number of segments used, performance peaks at 10, but increasing it above 3 does not result in significant performance difference. This means that the model does not necessarily create stronger temporal structure when provided with more than 3 parts of the video.

        \begin{table}[t!]
        \centering
        \caption{Mask Effect Results on Face Input}
 \begin{tabular}{cccc}
   \multicolumn{2}{c}{\includegraphics[width=0.2\linewidth]{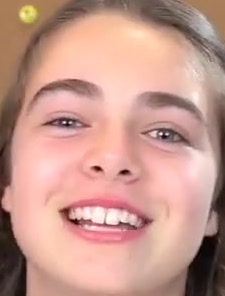}} &\multicolumn{2}{c}{\includegraphics[width=0.2\linewidth]{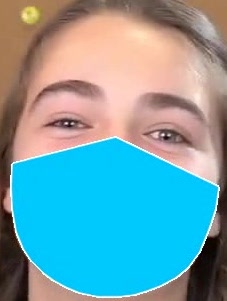}}\\
        \end{tabular}
        \begin{tabular}{|c|cc|c|}
        \hline
        \multirow{2}{*}{\textbf{Segments}} & \multicolumn{2}{c|}{\textbf{ROC AUC}} & \multirow{2}{*}{\textbf{Performance}}\\ \cline{2-3}
                                           & \multicolumn{1}{c|}{Default}  & Mask  &                 	                            \\ \hline		
        1                                  & \multicolumn{1}{c|}{0.755}    & 0.728 &   {${\color{red}-2.7\%}$}           \\ \hline        
        3                                  & \multicolumn{1}{c|}{0.769}    & 0.733 &  {${\color{red}-3.6\%}$}             \\ \hline			
        5                                  & \multicolumn{1}{c|}{0.767}    & 0.732 &  {${\color{red}-3.5\%}$}             \\ \hline				
        10                                 & \multicolumn{1}{c|}{0.770}    & 0.741 &  {${\color{red}-2.9\%}$}            \\ \hline                      
        \end{tabular}
        \label{effect1}
        \end{table}

        \begin{table}[t!]
        \centering
        \caption{Mask Effect Results on Full Body Input}
  \begin{tabular}{cccc}
   \multicolumn{2}{c}{\includegraphics[width=0.3275\linewidth]{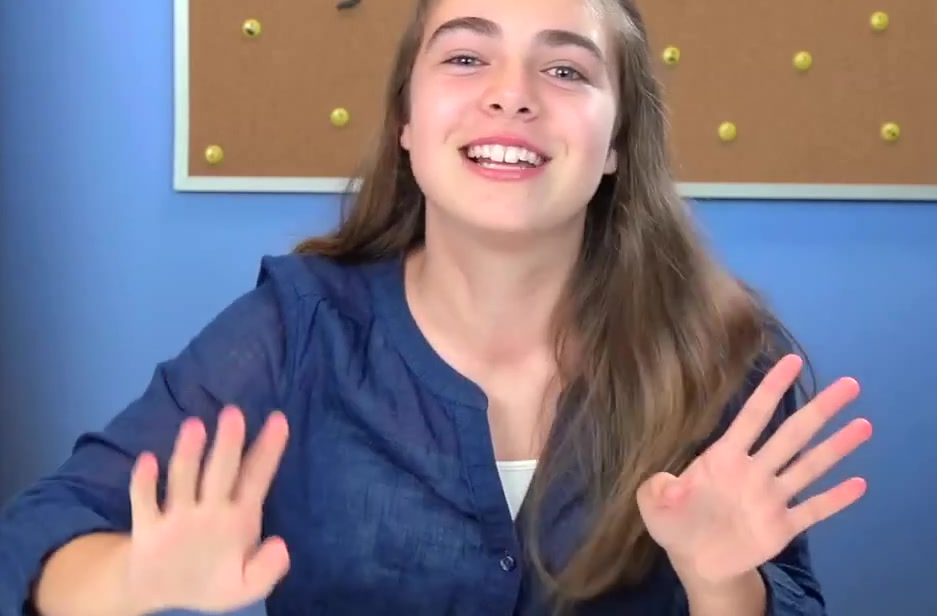}} &\multicolumn{2}{c}{\includegraphics[width=0.3275\linewidth]{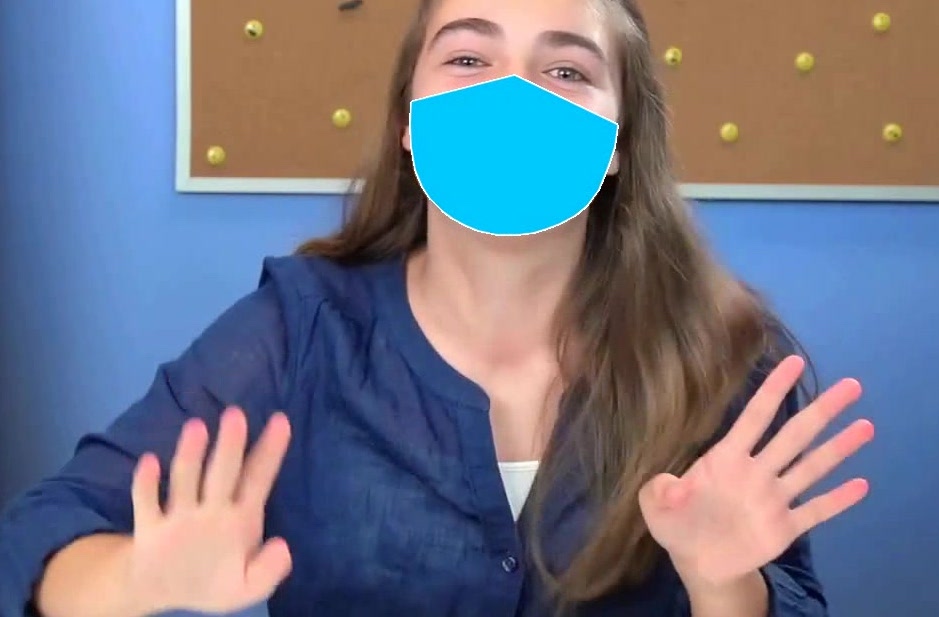}}\\
        \end{tabular}                                                                                   
        \begin{tabular}{|c|cc|c|}
        \hline
        \multirow{2}{*}{\textbf{Segments}} & \multicolumn{2}{c|}{\textbf{ROC AUC}} & \multirow{2}{*}{\textbf{Performance}}\\ \cline{2-3}
                                           & \multicolumn{1}{c|}{Default}  & Mask  &                                           \\ \hline 		
        1                                  & \multicolumn{1}{c|}{0.752}    & 0.752 &  -                                      \\ \hline 	
        3                                  & \multicolumn{1}{c|}{0.759}    & 0.758 &   {${\color{red}-0.1\%}$}        \\ \hline 		
        5                                  & \multicolumn{1}{c|}{0.758}    & 0.754 &  {${\color{red}-0.4\%}$}         \\ \hline 		
        10                                 & \multicolumn{1}{c|}{0.761}    & 0.759 &   {${\color{red}-0.2\%}$}       \\ \hline      
        \end{tabular}
        \label{effect2}
        \end{table}
        
        \subsubsection{Mask Effect on Full Body Input}
        Looking at Table \ref{effect2}, which shows results on full body input, the first and most important observation we make, is that performance decrease is very little to none (0-0.4\%). These results suggest that the model can exploit body information in such a way, that even with the application of a face mask, and consequently face information loss, it only suffers minimal performance drop. We also note the same pattern of performance with the masked face input results, which is better performance as complexity goes up. However, performance increase from 3 to 10 segments is minimal (0.1\%), which again suggests working towards the speed side of the trade-off.

	    \subsubsection{Masked Face vs Masked Full Body Results}
        Lastly, in Table \ref{effect3} we compare model performance with masked face versus masked full body crop, and show that incorporating the whole body in the input gives superior results over face crop. With black we highlight the best overall result, whereas with blue we highlight the result of the suggested optimal model, regarding the performance vs speed trade-off discussed earlier. The obvious conclusion is that moving towards bodily expression recognition is our best option, when the face is occluded. However, this is only a baseline result, which we could build on and pursue improvements by enhancing our model.

        \begin{table}[t!]
        \centering
        \caption{Masked Input Result Comparison}
\begin{tabular}{cccc}
        \multicolumn{2}{c}{\includegraphics[width=0.2\linewidth]{figures/face_mask}} &\multicolumn{2}{c}{\includegraphics[width=0.4\linewidth]{figures/body_mask}}\\
        \end{tabular}
        \begin{tabular}{|c|cc|c|}
        \hline
        \multirow{2}{*}{\textbf{Segments}} & \multicolumn{2}{c|}{\textbf{ROC AUC}} & \multirow{2}{*}{\textbf{Performance}}\\ \cline{2-3}
                                           & \multicolumn{1}{c|}{Masked Face}  & Masked Full Body  &                                            			       \\ \hline
        1                                  & \multicolumn{1}{c|}{0.728}    & 0.752 &   {${\color{green}+2.4\%}$}        \\ \hline
        3                                  & \multicolumn{1}{c|}{0.733}    & \color{blue} \textbf{0.758} &   {${\color{green}+2.5\%}$}         \\ \hline
        5                                  & \multicolumn{1}{c|}{0.732}    & 0.754 &  {${\color{green}+2.2\%}$}          \\ \hline
        10                                 & \multicolumn{1}{c|}{0.741}    & \textbf{0.759} &   {${\color{green}+1.8\%}$}            				  \\ \hline                         
        \end{tabular}
        \label{effect3}
        \end{table}

    \begin{figure}[t!]
	\centering
   	\includegraphics[width=1\linewidth]{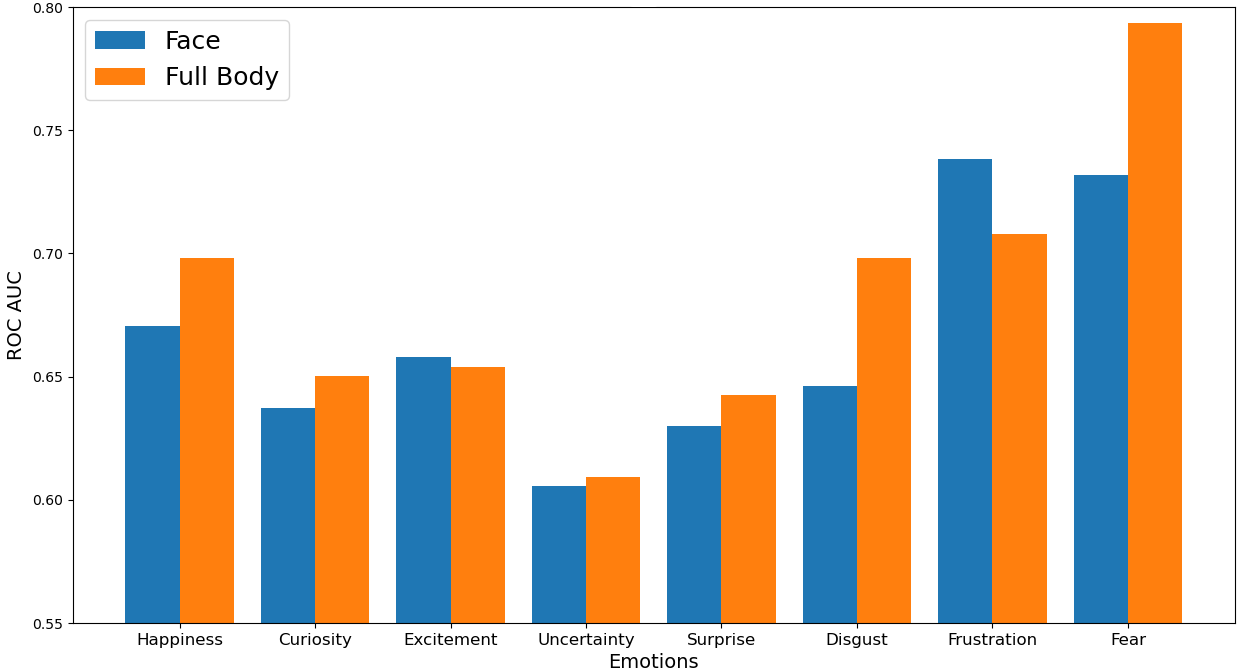}
	\caption{Masked Input Modality per Emotion Performance}
	\label{peremo}
    \end{figure}
    
    \subsection{Per Emotion Performance}
    In Fig. \ref{peremo}, we report per emotion ROC AUC and compare masked face versus masked full body input performance. Full body outperforms face in all emotions, except for Excitement and Frustration. This could be translated as these two emotions being expressed more by facial than bodily features from the children involved and incorporating the body in this case misleads the network. For Fear, performance is a lot higher with full body compared to face, which intuitively makes sense as children tend to utilize their body more to express fear \cite{F19}. Happiness is not conventionally an emotion with intense expressions, as most people think of just a simple smile, which is obstructed by the mask, but the model manages to recognize it at a decent level. Another conclusion we could come up to is that for some emotion pairs, like Curiosity-Uncertainty or Excitement-Surprise, which intuitively are quite similar to each other, performance might be lower for each emotion individually, because it is harder for the model to distinguish one from the other.

    \subsection{Visual Explanation}
    To have a better understanding of the mask effect on performance, we utilize a technique for producing visual explanations for predictions. We wish to explore where our model focuses in the input image and how its behaviour varies for the different emotion category targets. The method we choose is Grad-CAM \cite{gradcam}, which uses the gradients of an emotion target flowing into the final convolutional layer, to produce a coarse localization map that highlights the important regions in the image for predicting that particular emotion. We provide some example frames of the activation mapping, where model focus increases from blue to red.

    \begin{figure}[t!]
    \centering
    \begin{subfigure}[b]{0.225\linewidth}
        \includegraphics[width=1\linewidth]{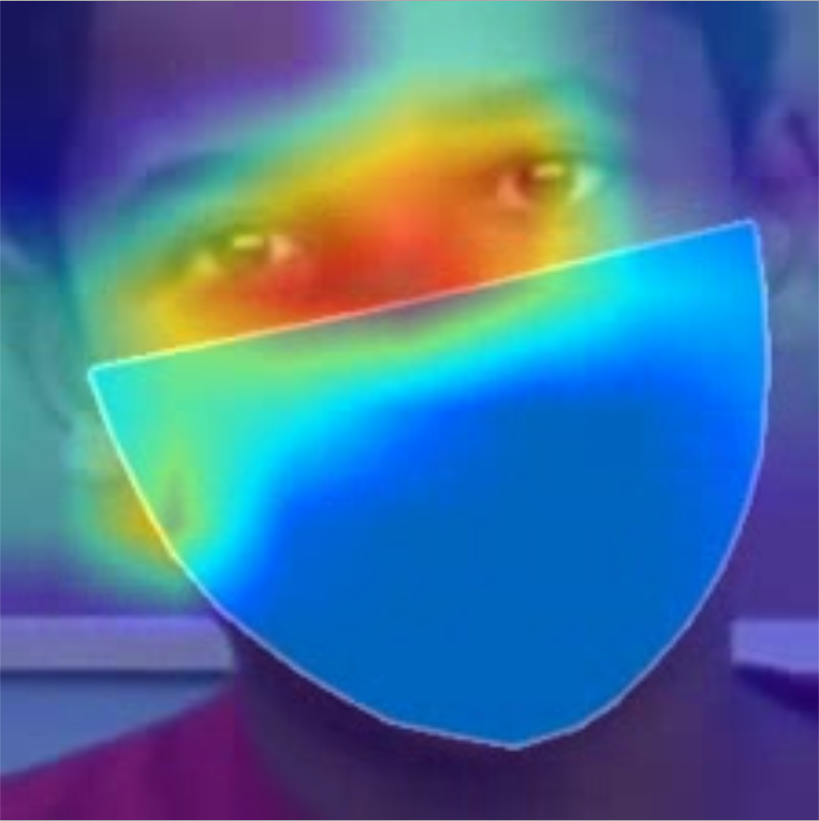}
        \caption*{Happiness}
    \end{subfigure}
    \begin{subfigure}[b]{0.225\linewidth}
        \includegraphics[width=1\linewidth]{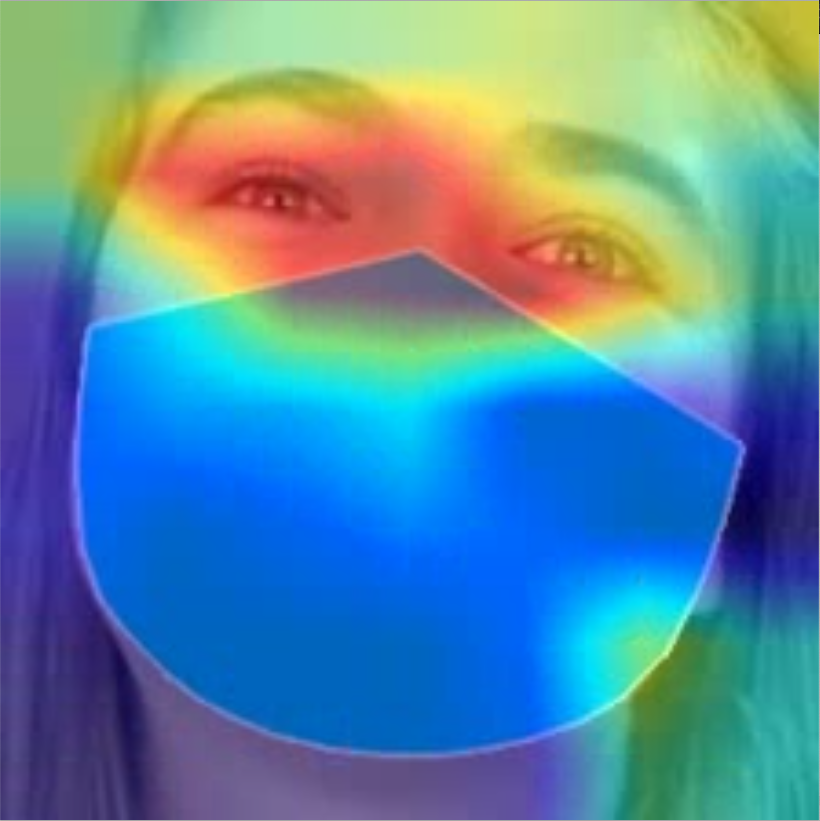}
        \caption*{Excitement}
    \end{subfigure}
    \begin{subfigure}[b]{0.225\linewidth}
        \includegraphics[width=1\linewidth]{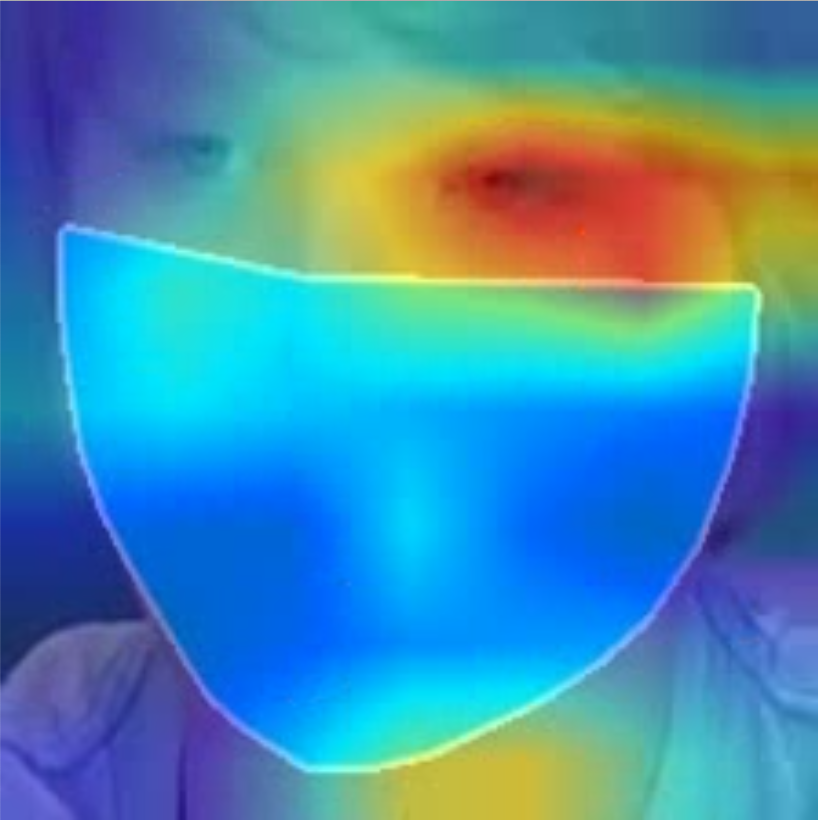}
        \caption*{Curiosity}
    \end{subfigure}
    \begin{subfigure}[b]{0.225\linewidth}
        \includegraphics[width=1\linewidth]{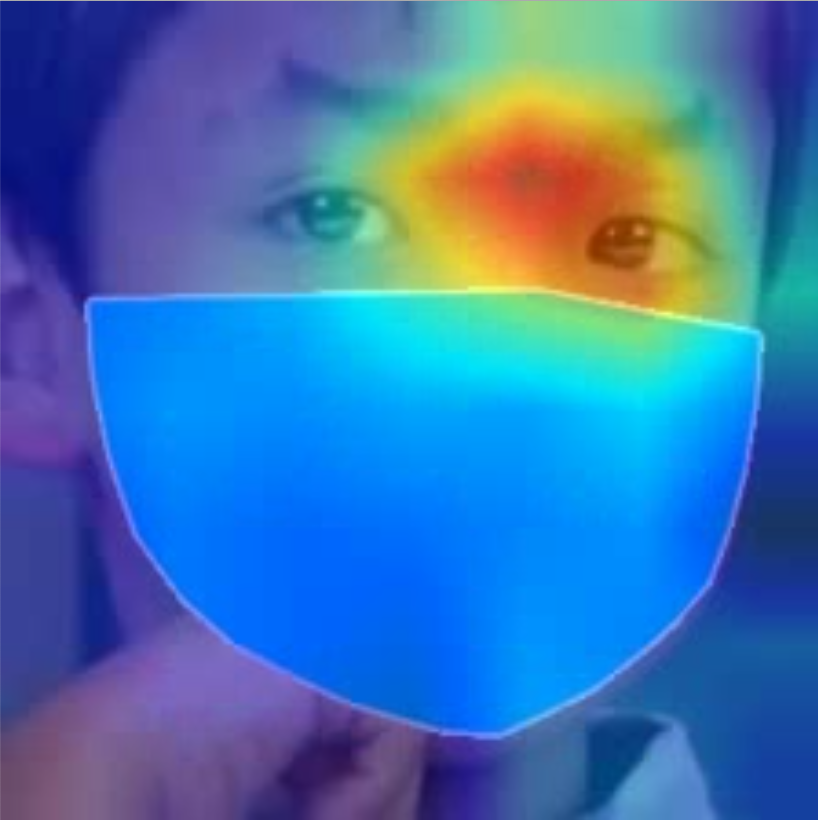}
        \caption*{Frustration}
    \end{subfigure}
    \caption{Face Decision Regions}
    \label{fac}
    \end{figure}

    \begin{figure}[t!]
    \centering
    \begin{subfigure}[b]{0.225\linewidth}
        \includegraphics[width=1\linewidth]{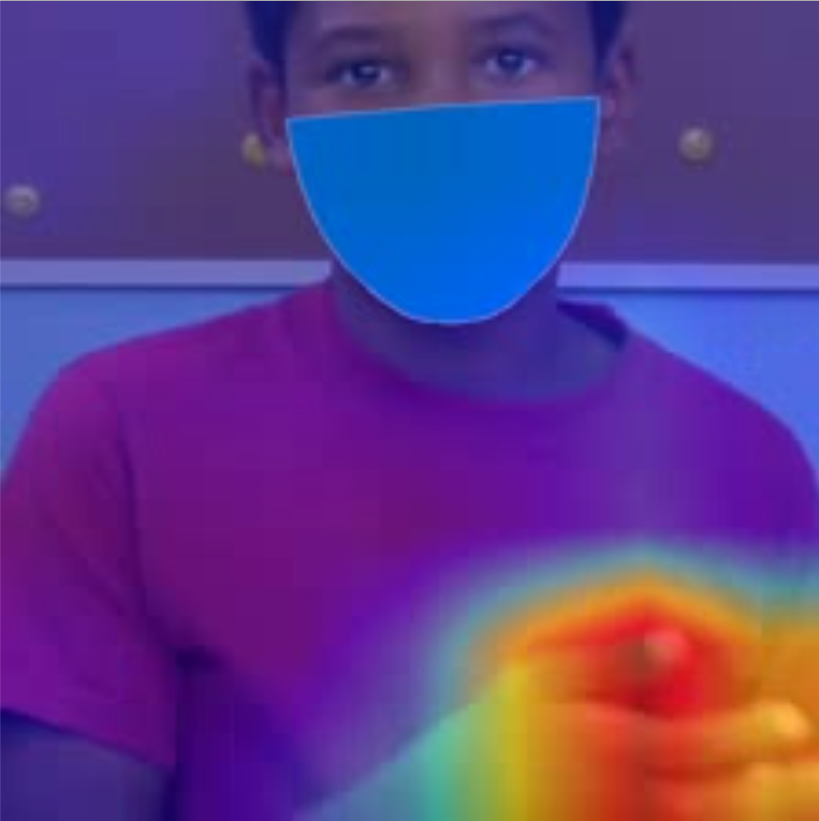}
        \caption*{Happiness}
    \end{subfigure}
    \begin{subfigure}[b]{0.225\linewidth}
        \includegraphics[width=1\linewidth]{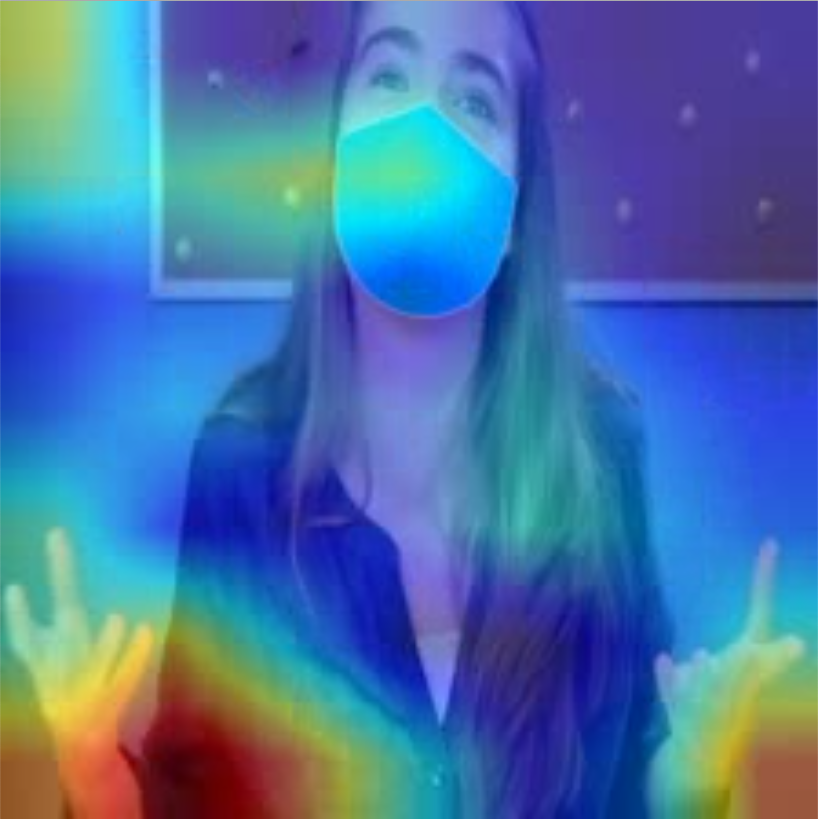}
        \caption*{Excitement}
    \end{subfigure}
    \begin{subfigure}[b]{0.225\linewidth}
        \includegraphics[width=1\linewidth]{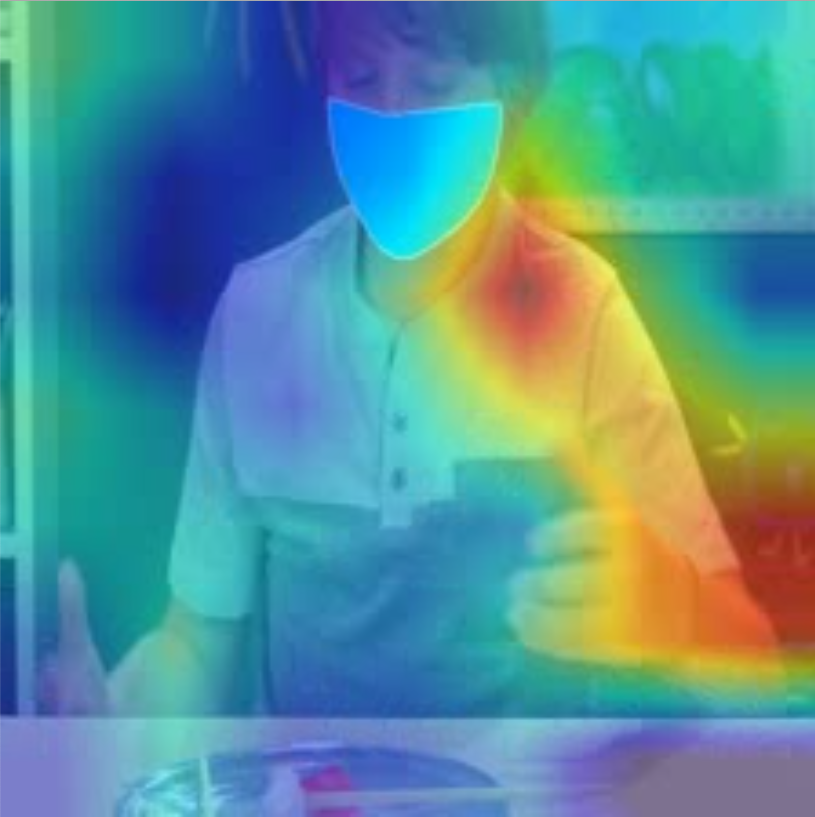}
        \caption*{Curiosity}
    \end{subfigure}
    \begin{subfigure}[b]{0.225\linewidth}
        \includegraphics[width=1\linewidth]{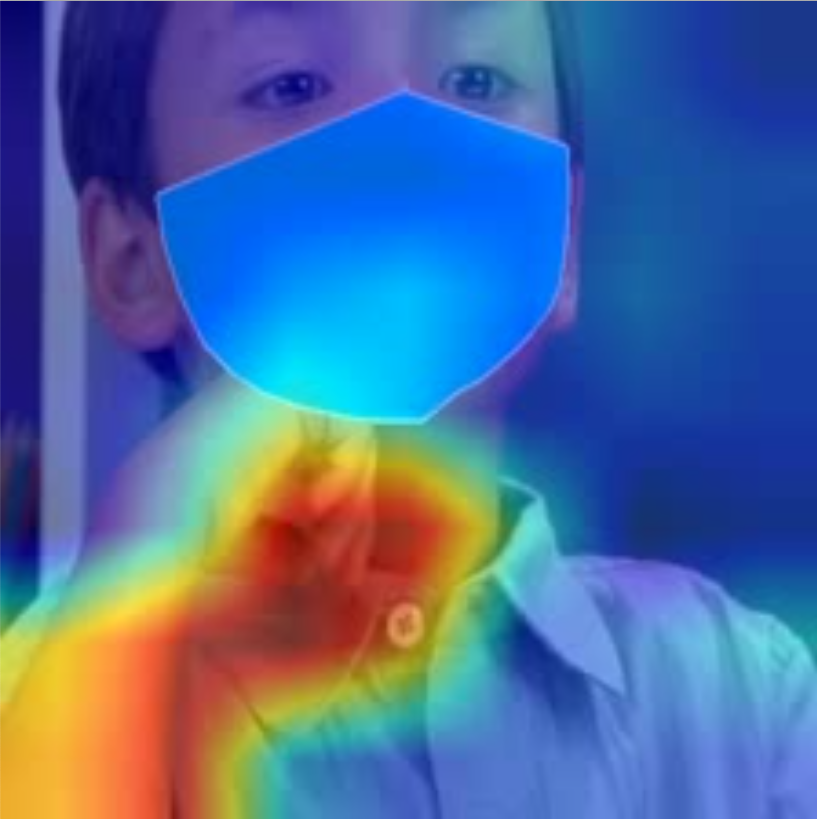}
        \caption*{Frustration}
    \end{subfigure}
    \caption{Body Decision Regions}
    \label{bod}
    \end{figure}

     \begin{figure}[t!]
    \centering
    \begin{subfigure}[b]{0.225\linewidth}
        \includegraphics[width=1\linewidth]{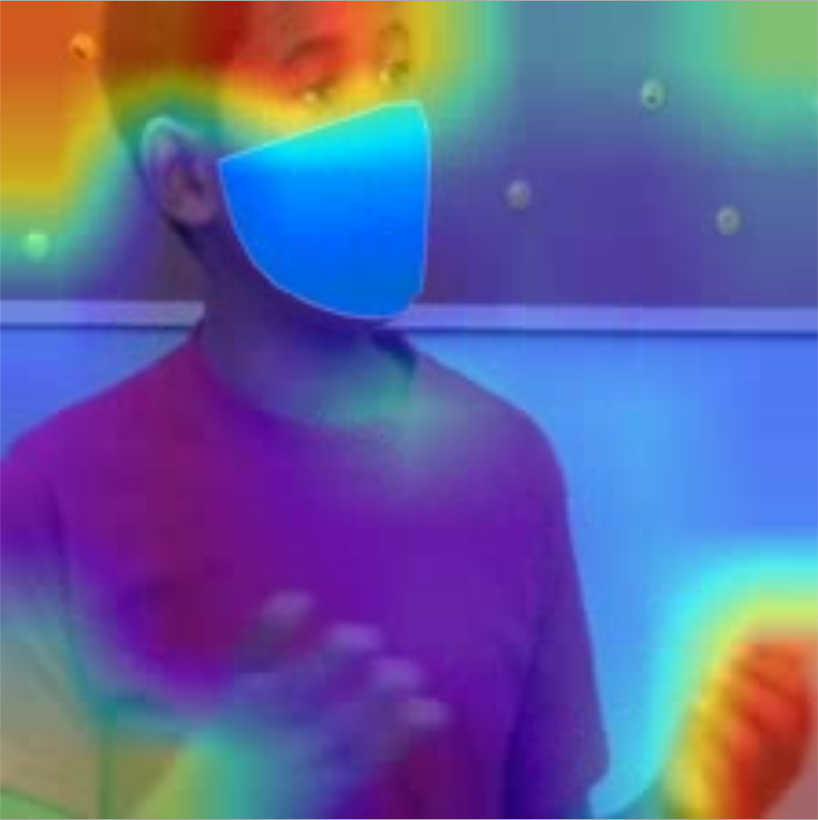}
        \caption*{Happiness}
    \end{subfigure}
    \begin{subfigure}[b]{0.225\linewidth}
        \includegraphics[width=1\linewidth]{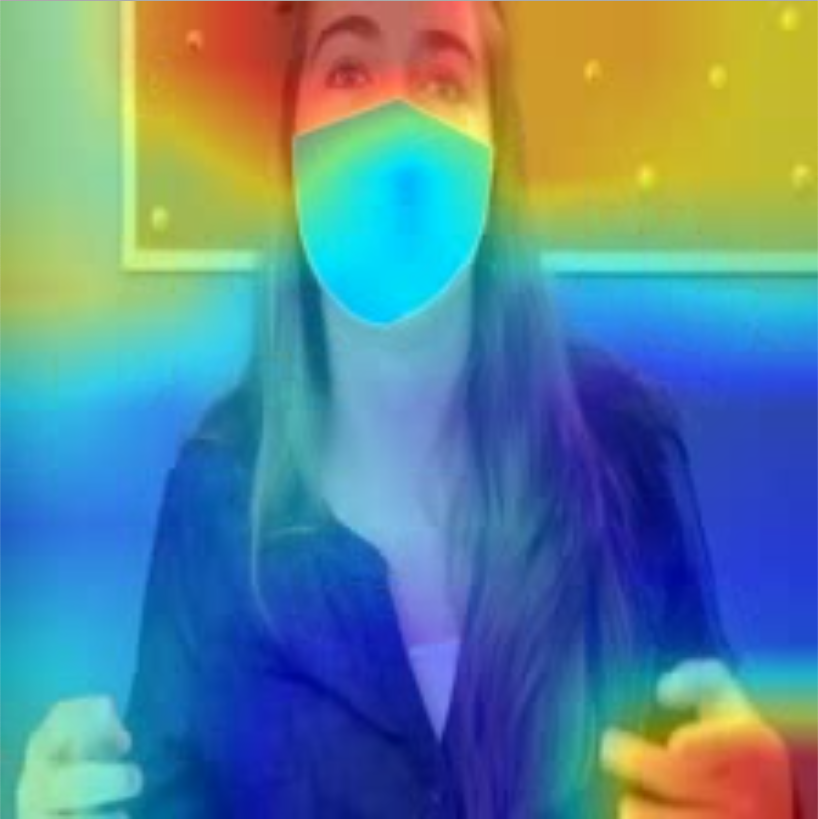}
        \caption*{Excitement}
    \end{subfigure}
    \begin{subfigure}[b]{0.225\linewidth}
        \includegraphics[width=1\linewidth]{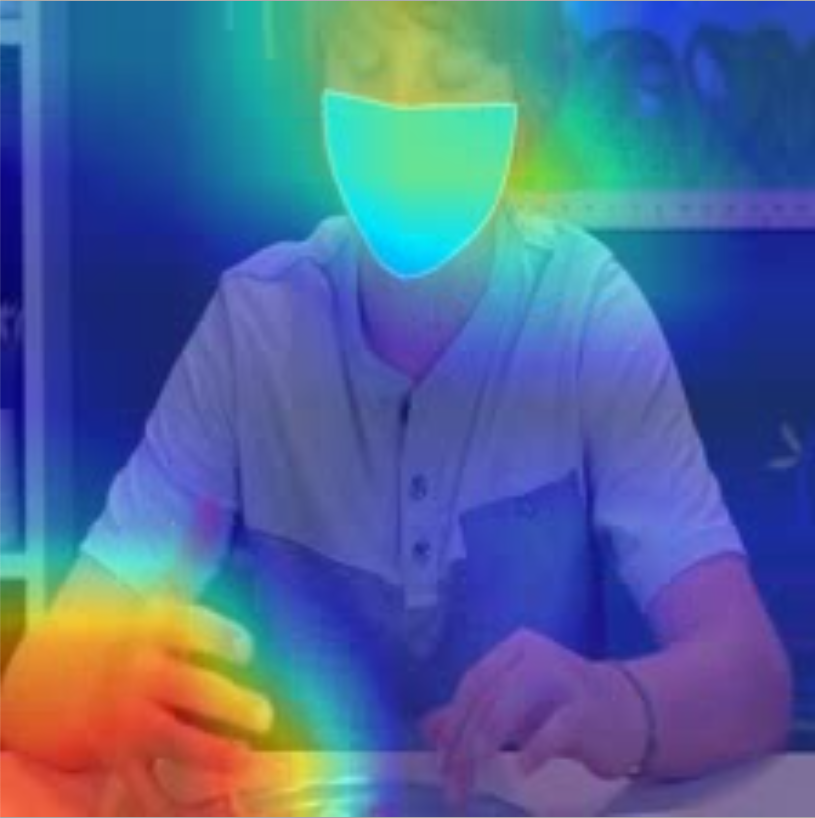}
        \caption*{Curiosity}
    \end{subfigure}
    \begin{subfigure}[b]{0.225\linewidth}
        \includegraphics[width=1\linewidth]{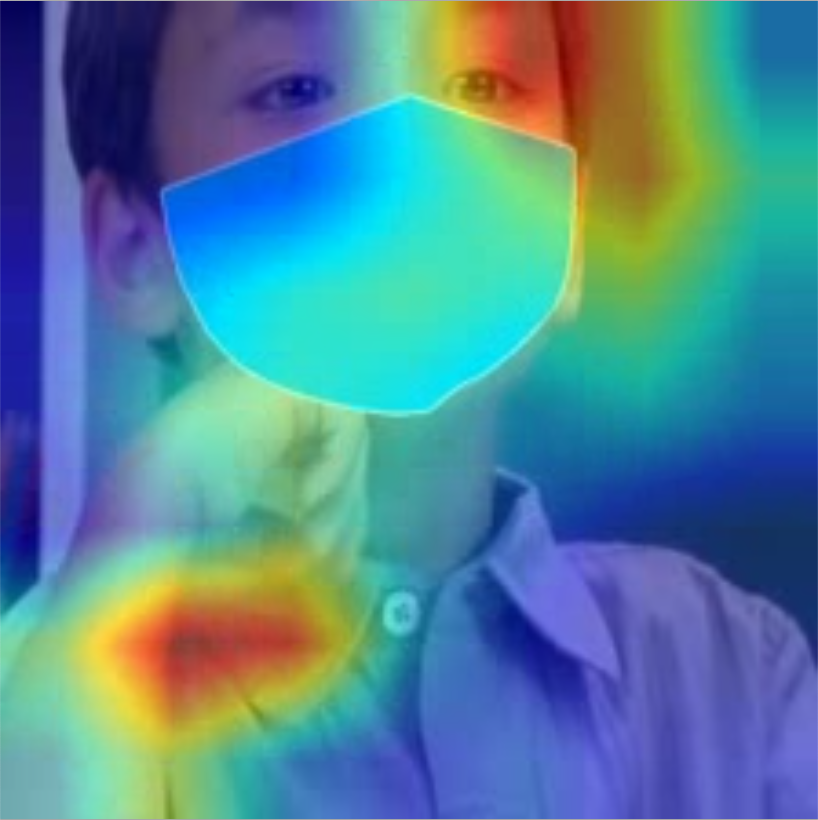}
        \caption*{Frustration}
    \end{subfigure}
    \caption{Mixed Decision Regions}
    \label{fus}
    \end{figure}
 
    Starting from the face examples (Fig. \ref{fac}), we can see that the model focuses on the upper part of the face. The facial features that could be utilized are the eyes (half-closed for Curiosity, wide open for Excitement), the eyebrows (raised for Excitement), and the forehead (frowning for Frustration).
    
    Regarding the body examples (Fig. \ref{bod}), the arms become visible and provide information that is utilized by the model. The bodily features that could be utilized are the hands (calm for Happiness, aroused for Excitement, investigating for Curiosity, fist for Frustration), the arms (wide open for Excitement), and the shoulders (shrugged for Curiosity).
    
    In Fig. \ref{fus}, we present several examples where the model focuses not only the body, but also on the face, fusing different modality information to make predictions. This suggests that it is able to learn both facial and bodily features in a single RGB stream.
    
    Overall, the model has learned to ignore noisy features, like the mask and the background. It is crucial to note, that the background is considered noise in this dataset, as the videos were recorded in a directed setup and it can be the same for different reaction topics. There is also large variability introduced by differences in the children's appearance due to clothing, body shape, size and hairstyles. These examples specify that the model is able to overcome these difficulties and focus on the expressive features.

    \subsection{Enhancement Results}
    We enhance the TSN-based model with the modality fusion and temporal modeling techniques and aspire to fully overcome the consequences of the face mask, by achieving performance as high as with the unmasked input.

        \subsubsection{Modality Fusion}
        In Table \ref{fustab}, we report fusion results after experimenting with two different aggregation functions: maximum and average. We also present an extra row of the plain body input, the performance of which is expected to be on the same scale with the masked face. A first observation we can make is, that using maximum as the aggregation function gives poor results, as it is actually outperformed by the plain body crop method. That might happen, because we are utilizing different modality information with a single input and wrong positive predictions (false positives) from one modality are possibly canceling out correct negative predictions (true negatives) from the other. On the other hand, averaging seems a choice that blends well, as it clearly improves performance. Intuitively, it makes sense to have a balanced consensus between modalities, as emotional expression cues can vary.
    
        \begin{table}[t!]
        \centering
        \caption{TSN Fusion Scheme Performance Comparison}
        \begin{tabular}{|c|c|c|}
        \hline
        \textbf{Input - 3 Segments}                      & \textbf{Aggregation} & \textbf{ROC AUC} \\ \hline \hline
        Masked Face                       &   -             							      & 0.733            \\ \hline \hline
        Plain Body                           &   -             							      & 0.736            \\ \hline \hline
        \multirow{2}{*}{Fusion} & Maximum             				      & 0.724             \\ \cline{2-3} 
                                                  &  Average                                  &  \textbf{0.764}           \\ \hline
        \end{tabular}
        \label{fustab}
        \end{table}

       \begin{table}[t!]
        \centering
        \caption{TSN vs TSM Model Performance Comparison}
        \begin{tabular}{|c|c|c|c|}
        \hline
        \textbf{Model}         & \textbf{Input - 3 Segments}  & \textbf{Shift}       & \textbf{ROC AUC} \\ \hline \hline
        TSN 							& Masked Face                                             & -                         &    0.733            \\ \hline \hline
        TSN                         & Masked Full Body                                       & -                         &    0.758                \\ \hline \hline
        \multirow{2}{*}{TSM}& \multirow{2}{*}{Masked Full Body}           & 1/8                      &    0.762                \\ \cline{3-4}
                                       &                                                     & 1/4                     &    \textbf{0.763}     \\ \hline
        \end{tabular}
        \label{tsmtab}
        \end{table}

        \subsubsection{Temporal Modeling}
        We experiment with the originally proposed channel shift fractions: 1/8 and 1/4. In Table \ref{tsmtab}, we observe that inserting TSM improves performance slightly. Either by shifting 1/4 or 1/8 of the channels, the difference is minimal. We come to the conclusion, that spatial feature learning plays a more important role for an emotional expression, while temporal structure is rather complementary.

        \subsubsection{Method Combination}
        In Table \ref{enhtab}, we report results when combining the TSM and fusion techniques. It seems that when utilizing both, the same conclusions as earlier apply. That means, TSM seems to give slight temporal modeling ability to the model and the fusion method results suggest that it effectively takes advantage of the face and body information separately, and possibly avoids irrelevant information confusion. The best overall performance is 0.768 ROC AUC and is achieved by the averaging fusion method, when using TSM with 1/4 partial shift. Compared to 0.769, which is the best face result achieved with no mask applied, reported in Table \ref{effect1}, we almost fully overcome face information loss and achieve similar performance. For reference, the last row reports the balanced ROC AUC average result from \cite{No+16}, where features extracted from \cite{openface} are used with an SVM, which our TSM Fusion method clearly outperforms.
    
        \begin{table}[t!]
        \centering
        \caption{Method Combination Performance Results}
        \begin{tabular}{|c|c|c|c|cc|}
        \hline
        \multirow{2}{*}{\textbf{Model}} & \multirow{2}{*}{\textbf{Input - 3 Seg.}} & \multirow{2}{*}{\textbf{Shift}} & \multirow{2}{*}{\textbf{Aggr.}} & \multicolumn{2}{c|}{\textbf{ROC AUC}}\\ \cline{5-6}
                                           &                                  &             &   &  \multicolumn{1}{c|}{Unb.}  & Bal.\\ \hline 
        TSN                         &Masked Face                   & -                             &    -                            & \multicolumn{1}{c|}{0.733}  & -          \\ \hline \hline
        TSN                         &Masked Full Body            & -                             &   -                             & \multicolumn{1}{c|}{0.758}  & -          \\ \hline \hline
        \multirow{4}{*}{TSM}& \multirow{4}{*}{Fusion} & \multirow{2}{*}{1/8} & Max.                  & \multicolumn{1}{c|}{0.729}     & -          \\ \cline{4-6} 
                                      &                                      &                                 & Avg.                     & \multicolumn{1}{c|}{0.767}     & -        \\ \cline{3-6} 
                                      &                                      & \multirow{2}{*}{1/4} & Max.                   & \multicolumn{1}{c|}{0.731}    & -         \\ \cline{4-6} 
                                      &                                      &                                 & Avg.                      & \multicolumn{1}{c|}{\textbf{0.768}} &  \textbf{0.696} \\ \hline \hline                        
        
        TSN                    & \multirow{2}{*}{Unmasked Face}& -       & -      & \multicolumn{1}{c|}{0.769}  & \multicolumn{1}{c|}{0.698} \\ \cline{1-1}\cline{3-6}
        \cite{No+16}            &              & -          & -       & \multicolumn{1}{c|}{-} & 0.620 \\ \hline
        
        \end{tabular}
        \label{enhtab}
        \end{table}

\section{Conclusion}
\label{section:conc}
In this work, we studied the effect of face occlusion on a CRI visual emotion recognition problem. In the presence of a face mask, performance from just the face drops considerably and urges us to incorporate the body modality. By providing the full body image as input, the model can sustain its performance and outperform the masked face case. Spatial information can be instrumental and yield great results, while temporal structure complements fittingly, as the consensus of several video segments provides additional emotional expression information. When enhancing the baseline model with temporal modeling and more importantly modality fusion, we almost fully overcome face information loss and achieve performance similar to the unmasked input case. Our visualizations provided insights suggesting a single RGB stream can ignore noise and learn both from facial, as well as bodily expressive features. An emotion recognition system with these capabilities can effectively tackle face occlusion forced by health and safety protocols, and be a core part of various applications in crucial areas like education and health care, for both adults and children.

\begin{acks}
This work has received funding from the European Union’s Horizon Europe research and innovation programme under grant agreement no. 101070381 (project: PILLAR-Robots).
\end{acks}

\bibliographystyle{ACM-Reference-Format}
\bibliography{ref.bib}

\end{document}